\documentclass[conference]{IEEEtran}
\IEEEoverridecommandlockouts
\usepackage{cite}
\usepackage{amsmath,amssymb,amsfonts}
\usepackage{algorithmic}
\usepackage{graphicx}
\usepackage{float}
\usepackage{textcomp}
\usepackage{xcolor}
\usepackage{booktabs}
\usepackage[labelsep=colon]{caption}
\def\BibTeX{{\rm B\kern-.05em{\sc i\kern-.025em b}\kern-.08em
    T\kern-.1667em\lower.7ex\hbox{E}\kern-.125emX}}
\begin{document}

\title{WHTMix: Efficient Stereo Depth Estimation via Walsh-Hadamard Token Mixing}

\author{
    \textit{Prathyush Sajith$^{1}$, Emadeldeen Hamdan$^{1}$, Ahmet Enis Cetin$^{1}$} \\[1ex]
    $^{1}$Dept. of Electrical and Computer Engineering, University of Illinois at Chicago \\
    Chicago, IL, USA
}

\maketitle

\begin{abstract}
Stereo depth estimation for driving, robotics, and augmented reality must run at high
resolution under tight latency budgets, yet in transformer-based matchers the global
self-attention that aggregates scene context grows quadratically with the number of
pixels and comes to dominate runtime. We show that the joint self-attention stage of a stereo
transformer, whose role is to spread context across both views, can be replaced by
a data-independent Walsh-Hadamard token mixer that mixes tokens globally in the
transform domain at log-linear cost, while the data-dependent cross-attention that
performs left-right correspondence is retained. On dense ground-truth driving data the mixer
matches the attention baseline in pre-training end-point error while reducing model compute by a
factor of 2.46 and stereo-pair inference latency by a factor of 2.65. A complexity
analysis shows the benefit is governed by the ratio of sequence length to channel
width, which explains why high-resolution stereo matching is a particularly
favorable setting and why classification transformers are not; we confirm this
token-to-channel scaling on non-stereo long-sequence benchmarks. We additionally investigate a hybrid log-disparity loss that up-weights small-disparity pixels corresponding to long-range objects, reducing distant-object error at no additional computational cost.
\end{abstract}


\section{Introduction}
Stereo depth estimation recovers a dense disparity map from a rectified image pair
and underpins robotics, autonomous driving and augmented reality. Transformer-based
matchers such as STTR~\cite{sttr} attain competitive accuracy by replacing a fixed
disparity search with attention, but their global self-attention scales as
$O(N^2)$ in the number of feature tokens $N$, making them slow at the high
resolutions these applications demand.

Attention plays two distinct roles inside a stereo transformer layer. A
\emph{joint self-attention} first spreads global context within and across the two
views; a subsequent \emph{cross-attention} performs the actual left-right
correspondence. Only the second is inherently content-dependent: matching a pixel to
its epipolar counterpart requires comparing image content. Context aggregation, by
contrast, can be performed by a fixed, data-independent operator. This motivates
replacing the joint self-attention with a cheap spectral token mixer while keeping
cross-attention for matching.

We use the orthonormal Walsh-Hadamard transform (WHT), a multiply-free,
real-valued relative of the Fourier transform computable in $O(N\log N)$ by a
butterfly. Our mixer, \textbf{WHTMix}, applies learnable per-frequency gains in the
WHT domain over both channels and tokens, giving a global receptive field at
log-linear cost. Our contributions are:
\begin{itemize}
\item A drop-in Walsh-Hadamard token mixer that replaces the joint self-attention of
a stereo transformer, cutting model compute $2.46\times$ and latency $2.65\times$ at
matched accuracy on synthetic data.
\item A complexity analysis giving a closed-form speedup of $2 + N/C$, with a
crossover at $N=2C$ and an end-to-end ceiling near $3\times$, that predicts exactly
where the mixer helps, matches measured latency, and is confirmed on non-stereo
Long-Range Arena tasks.
\end{itemize}
Beyond these, we report two additional investigations of the resulting model: a
hybrid log-disparity loss that substanstially reduces far-object EPE at no latency cost, and an
honest account of the synthetic-to-real gap a fixed mixer incurs when real training
data are scarce.

\section{Related Work}
\textbf{Stereo matching.} Cost-volume networks with 3D
convolutions~\cite{psmnet,gwcnet} and iterative refinement methods such as
RAFT-Stereo~\cite{raftstereo}, CREStereo~\cite{crestereo} and IGEV-Stereo~\cite{igev}
dominate standard benchmarks; unified matching architectures~\cite{unimatch} and
adaptive frequency selection~\cite{selstereo} extends the same paradigm, while
foundation models push zero-shot generalization~\cite{foundstereo}.
STTR~\cite{sttr} recast matching as a sequence-to-sequence problem using alternating
self- and cross-attention along epipolar lines; the same division of labour underlies
detector-free feature matching~\cite{loftr}. Closest to us, HART~\cite{hart} builds
a stereo transformer on Hadamard-\emph{product} attention; we instead replace the joint
self-attention with the Walsh-Hadamard \emph{transform} \cite{agaian1985hadamard,agaian2011hadamard} as a fixed token mixer, keeping
dot-product cross-attention for matching. We adopt an STTR-style backbone as our testbed
because it is the paradigm in which global self-attention is the bottleneck.

\textbf{Efficient and spectral token mixers.} Dot-product self-attention~\cite{transformer}
is quadratic in sequence length; efficient variants use windowed
attention~\cite{swin}, low-rank or kernelized
approximations~\cite{linformer,performer,nystromformer}, implicit long
convolutions~\cite{hyena}, or selective state spaces~\cite{mamba}, while
FlashAttention~\cite{flashattn,flashattn2} lowers attention's \emph{memory} but not its
quadratic compute. Spectral mixers are a distinct line: FNet~\cite{fnet} replaces attention with
an unparameterized 2D Fourier transform; GFNet~\cite{gfnet} learns global filters in the
Fourier domain; AFNO~\cite{afno} adds a channel mixer in frequency space; and
Walsh-Hadamard layers have cut the convolution and projection cost in deep
networks~\cite{panfwht,panblock,whtnn,zhu2026vhu,pan2023hybrid}. Architectural studies further suggest that the
surrounding block matters more than the identity of the mixer~\cite{mlpmixer,metaformer},
which is what licenses do, substituting a cheaper operator wherever content dependence is not
required. Our mixer is closest to GFNet but uses the real,
multiply-free Walsh-Hadamard transform and, unlike these classification-oriented works,
is applied selectively to the context-mixing stage of a dense correspondence model while
retaining data-dependent cross-attention.

\section{Method}
\subsection{Backbone}
We follow an STTR-inspired pipeline: a shared CNN feature extractor downsamples each
image by $8\times$; a fixed 2D sinusoidal positional encoding is added; a stack of
$L$ transformer layers alternate joint self-attention (over the concatenated
left-right tokens, length $N=2\,(H/8)(W/8)$) with two cross-attentions; a
normalized correlation volume and a soft-argmin regression head produce the
disparity. Channel width is $C$. We replace only the joint self-attention.

\subsection{Walsh-Hadamard Token Mixer}
\begin{figure}[t]
\centerline{\includegraphics[width=0.47\textwidth]{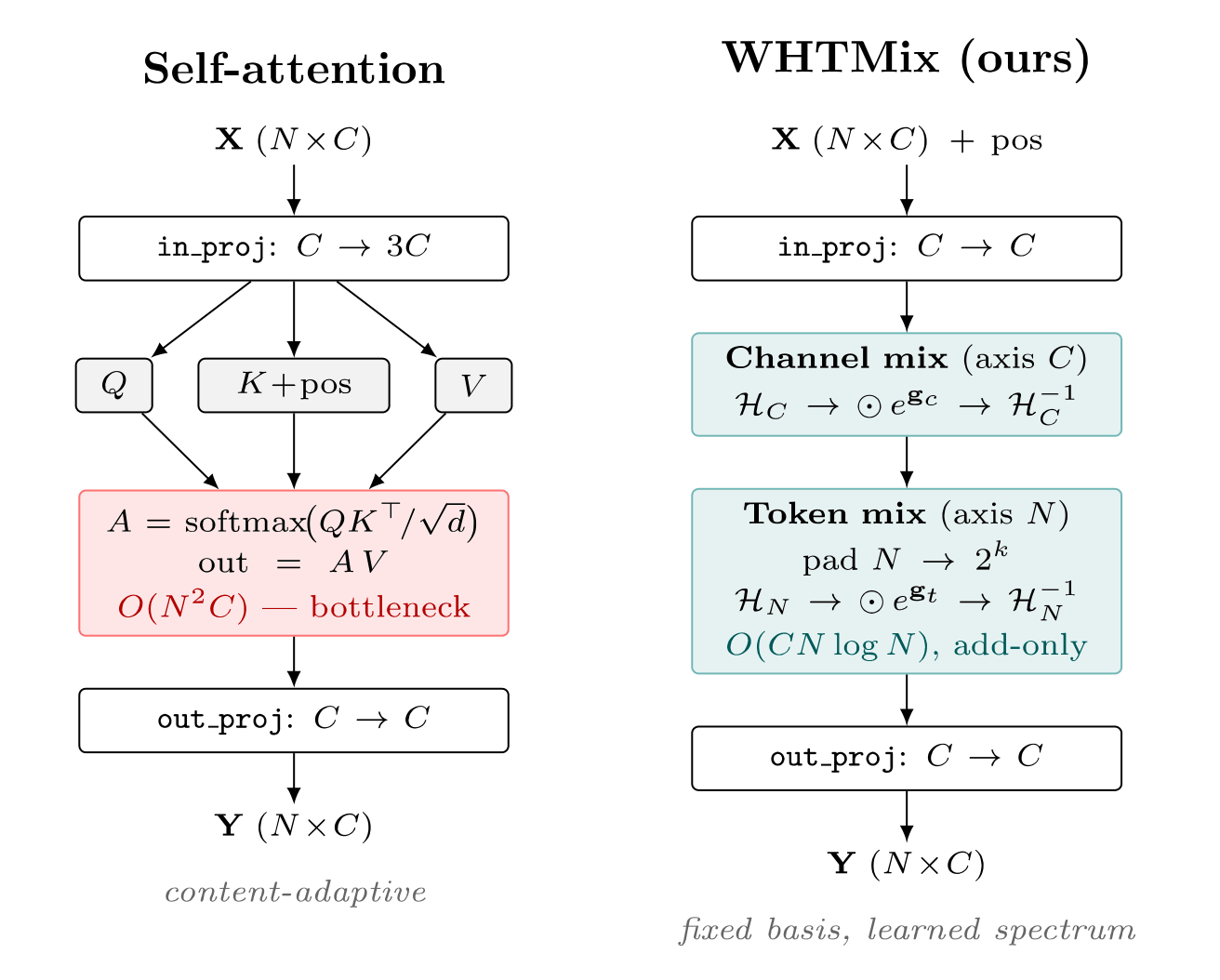}}
\caption{Architectural difference between self-attention and WHTMix}
\label{fig:arch}
\end{figure}

Let $\mathbf{X}\in\mathbb{R}^{N\times C}$ be the joint token sequence and let
$\mathcal{H}$ denotes the orthonormal WHT (the Sylvester-Hadamard matrix normalized by
$\sqrt{n}$, defined for $n$ a power of two and applied by an $O(n\log n)$ butterfly).
WHTMix computes
\begin{align}
\mathbf{Z} &= \mathbf{X}\mathbf{W}_{\mathrm{in}}, \\
\mathbf{Z} &\leftarrow \mathcal{H}_C^{-1}\!\big(e^{\mathbf{g}_c}\odot \mathcal{H}_C(\mathbf{Z})\big), \\
\mathbf{Z} &\leftarrow \mathcal{H}_N^{-1}\!\big(e^{\mathbf{g}_t}\odot \mathcal{H}_N(\mathbf{Z})\big), \\
\mathbf{Y} &= \mathbf{Z}\mathbf{W}_{\mathrm{out}},
\end{align}
where $\mathcal{H}_C$ and $\mathcal{H}_N$ act over channels and tokens respectively,
and $\mathbf{g}_c,\mathbf{g}_t$ are learnable log-domain per-frequency gains
(initialized to zero, i.e., unit gain). The token axis is zero-padded to the next
power of two before $\mathcal{H}_N$. Both transforms are orthonormal, hence
magnitude preserving, so layers stack stably; as in FNet/GFNet, the block is linear
and the layer feed-forward network supplies the nonlinearity. WHTMix mirrors the
attention call signature and is a literal drop-in for the self-attention module.
Fig.~\ref{fig:arch} contrasts the two token mixers.

\subsection{Hybrid Log-Disparity Loss}
Standard masked $\ell_1$ is dominated by near objects with large disparities, so far
objects (with small disparities) are unsupervised. We weight each valid pixel $p$ by
\begin{equation}
w_p = \mathrm{clip}\!\left(\tau / d_p,\; 1,\; w_{\max}\right),
\end{equation}
As in focal loss~\cite{focalloss}, this reweights the existing pixels rather than
resampling them, so it costs nothing at inference. With ground-truth disparity $d_p$,
pixels below $\tau$ px are up-weighted up to
$w_{\max}$, and minimize $\frac{1}{\sum_p w_p}\sum_{p} w_p\,|d_p-\hat d_p|$. We use
$\tau=16$, $w_{\max}=32$.

\subsection{Complexity Analysis}
Counting multiply-accumulates (MACs; FLOPs $=2\times$MACs), a self-attention block
costs $4NC^2$ (projections) $+\,2N^2C$ (the quadratic core), whereas WHTMix costs
$\approx 2NC^2$ (two projections; the butterfly is add-only, contributing no MACs).
The token-mixer speedup is therefore
\begin{equation}
S_{\mathrm{mod}} = \frac{4NC^2 + 2N^2C}{2NC^2} = 2 + \frac{N}{C},
\label{eq:speedup}
\end{equation}
so the gain is governed by the token-to-channel ratio, with a crossover at $N=2C$
below which attention is projection-bound. Our stereo model at $512\times1024$
($N=16{,}384$, $C=64$) gives $N/C=256$; a classification ViT~\cite{vit}
($N\!\approx\!196$, $C\!=\!768$) gives $0.26$ and no benefit. Because the retained
cross-attentions (each on $N/2$ tokens) and the feed-forward network are shared, the
\emph{end-to-end} layer speedup is capped:
\begin{equation}
S_{\mathrm{e2e}} = \frac{16NC^2+3N^2C}{14NC^2+N^2C}
= \frac{16C+3N}{14C+N}\;\xrightarrow[N\to\infty]{}\;3.
\label{eq:ceiling}
\end{equation}

\begin{figure*}[t]
\centerline{\includegraphics[width=0.98\textwidth]{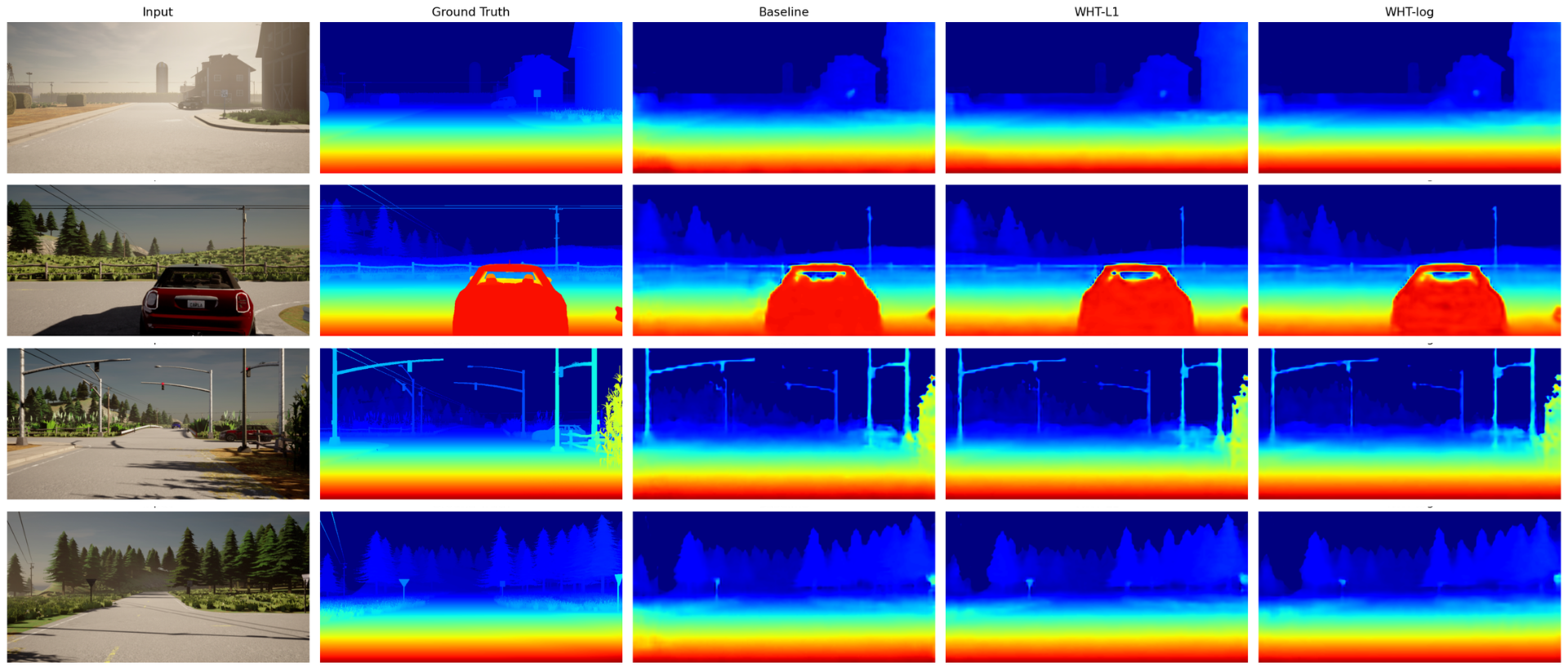}}
\caption{Qualitative comparison on CARLA (input, ground truth, and the three
models). WHTMix produces disparity maps visually similar from the
attention baseline; the log-disparity variant sharpens distant structure.}
\label{fig:qual}
\end{figure*}
\vspace{2mm}
\section{Experiments}
\subsection{Setup}
We pre-train on a synthetic
CARLA~\cite{carla} stereo set (about 2{,}214 triplets,
90/10 split) with AdamW ($\mathrm{lr}=10^{-4}$, weight decay $0.01$),
batch size 32, bf16 autocast, at $512\times1024$, and fine-tune on KITTI
2015~\cite{kitti} at $384\times1248$ with a frozen-backbone warm-up.
All models use $C=64$, $8$ heads, $L=3$ layers, maximum disparity $256$. We compare
three variants that differ by one factor at a time: \textbf{Baseline} (softmax
self-attention), \textbf{WHTMix (L1)} (our mixer, $\ell_1$ loss) and
\textbf{WHTMix (log)} (our mixer, log-disparity loss). We report end-point error (EPE) over valid pixels and a
per-disparity-bin breakdown, parameters, GFLOPs measured with fvcore, and batch-1 latency on an RTX 4090, reported as
the mean of $50$ timed iterations after $10$ untimed warm-up iterations, with
\texttt{cuda.synchronize()} before and after each timed call. Baseline attention runs
through PyTorch's fused scaled-dot-product attention (memory-efficient backend at
fp32 inference), so the speedup is measured against an optimized kernel rather than a
naive implementation. Evaluation uses a held-out CARLA split of
$246$ pairs and the $200$ KITTI pairs. All runs are single-seed, the efficiency figures
(parameters, GFLOPs, latency) are deterministic given the architecture and are
unaffected.

\subsection{Accuracy and Efficiency}
Table~\ref{tab:main} reports the main results. On CARLA, WHTMix matches the
attention baseline in EPE ($2.27$ vs.\ $2.29$; $95\%$ CI $[-0.04,+0.08]$, so the two are
statistically indistinguishable) while using
$2.46\times$ fewer FLOPs and running $2.65\times$ faster. On sparse ground-truth KITTI data, with
only 200 ; the fixed mixer is less accurate than
adaptive attention ($2.670$ vs.\ $2.204$~px): attention recomputes its mixing weights from each input, whereas the WHTMix mixing
pattern is fixed after training and can adapt only through its per-frequency gains.
Efficiency gains are essentially identical in both domains. WHTMix also uses slightly
fewer parameters ($0.759$ vs.\ $0.783$~M); the rest of the network is identical across variants.

\begin{table}[H]
\caption{Accuracy and efficiency on CARLA and KITTI. EPE is pixel-weighted over all
valid pixels; $\pm$ is the standard deviation of per-frame EPE across the test set
(scene-to-scene spread).}
\label{tab:main}
\centering
\footnotesize
\setlength{\tabcolsep}{4pt}
\begin{tabular}{lcccc}
\toprule
Model & Params\,(M) & GFLOPs & Lat.\,(ms) & EPE\,(px) \\
\midrule
\multicolumn{5}{c}{\textit{CARLA ($512\times1024$)}}\\
Baseline (attn)   & 0.783 & 348.9 & 95.7 & $2.294 \pm 1.218$ \\
WHTMix (L1)       & \textbf{0.759} & \textbf{141.9} & \textbf{36.0} & $\mathbf{2.274 \pm 1.145}$ \\
\midrule
\multicolumn{5}{c}{\textit{KITTI 2015 ($384\times1248$)}}\\
Baseline (attn)   & 0.783 & 294.6 & 81.8 & $\mathbf{2.204 \pm 1.166}$ \\
WHTMix (L1)       & \textbf{0.759} & \textbf{121.6} & \textbf{31.8} & $2.670 \pm 1.585$ \\
\bottomrule
\end{tabular}
\end{table}

Table~\ref{tab:bins} breaks EPE down by disparity. The log-disparity loss cuts error
on the far, small-disparity bins significantly (2--4~px: $3.29\!\to\!1.60$;
4--8~px: $3.36\!\to\!1.94$) at the cost of a modest overall regression. The trade is
systematic rather than incidental: against WHTMix (L1) every bin below $16$~px
improves and every bin above it degrades (32--64~px: $3.62\!\to\!4.01$), so the loss
reallocates supervision across the disparity range at unchanged parameters and
latency.

\begin{table}[H]
\caption{Per-disparity-bin EPE (px) on CARLA. The smaller the bins, the farther the objects.}
\label{tab:bins}
\centering
\setlength{\tabcolsep}{3.2pt}
\footnotesize
\begin{tabular}{lcccccccc}
\toprule
Model & 0--2 & 2--4 & 4--8 & 8--16 & 16--32 & 32--64 & 64--128 & 128$+$ \\
\midrule
Baseline    & 0.64 & 2.96 & 3.04 & 2.89 & \textbf{2.79} & 3.62 & 3.12 & 1.70 \\
WHTMix (L1) & 0.57 & 3.29 & 3.36 & 2.94 & 2.91 & \textbf{3.62} & \textbf{3.04} & \textbf{1.68} \\
WHTMix (log)& \textbf{0.56} & \textbf{1.60} & \textbf{1.94} & \textbf{2.49} & 3.03 & 4.01 & 3.44 & 1.96 \\
\bottomrule
\end{tabular}
\end{table}

\subsection{Latency Scaling}
\begin{figure}[H]
\centerline{\includegraphics[width=0.48\textwidth]{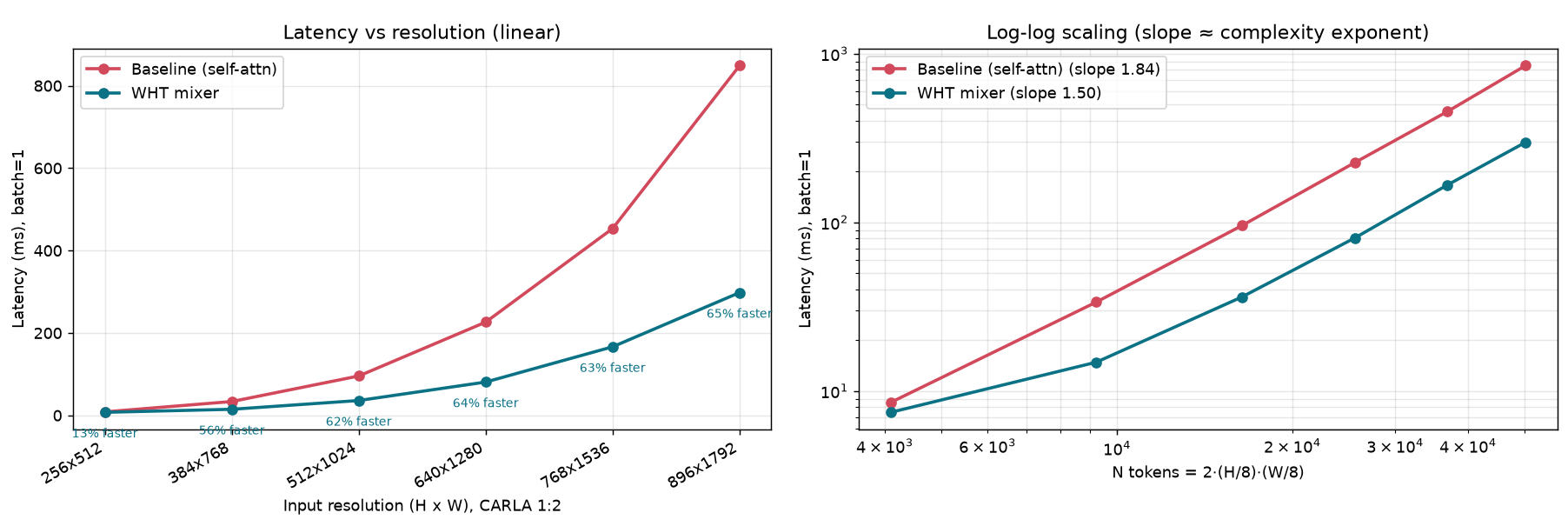}}
\caption{End-to-end batch-1 latency of the full stereo model vs.\ input resolution (CARLA $1{:}2$ aspect, RTX 4090). \emph{Left}: linear axes, each point annotated with the latency WHTMix saves over the attention baseline, growing from $13\%$ ($1.15\times$) at $256\times512$ to $65\%$ ($2.85\times$) at $896\times1792$. \emph{Right}: the same runs on log-log axes against the joint token count $N=2(H/8)(W/8)$; the fitted slopes ($1.84$ baseline vs.\ $1.50$ WHTMix) estimate the complexity exponent and stay linear for both models because the retained cross-attention keeps an $N^2C$ term.}
\label{fig:latency}
\end{figure}

\begin{figure}[H]
\centerline{\includegraphics[width=0.48\textwidth]{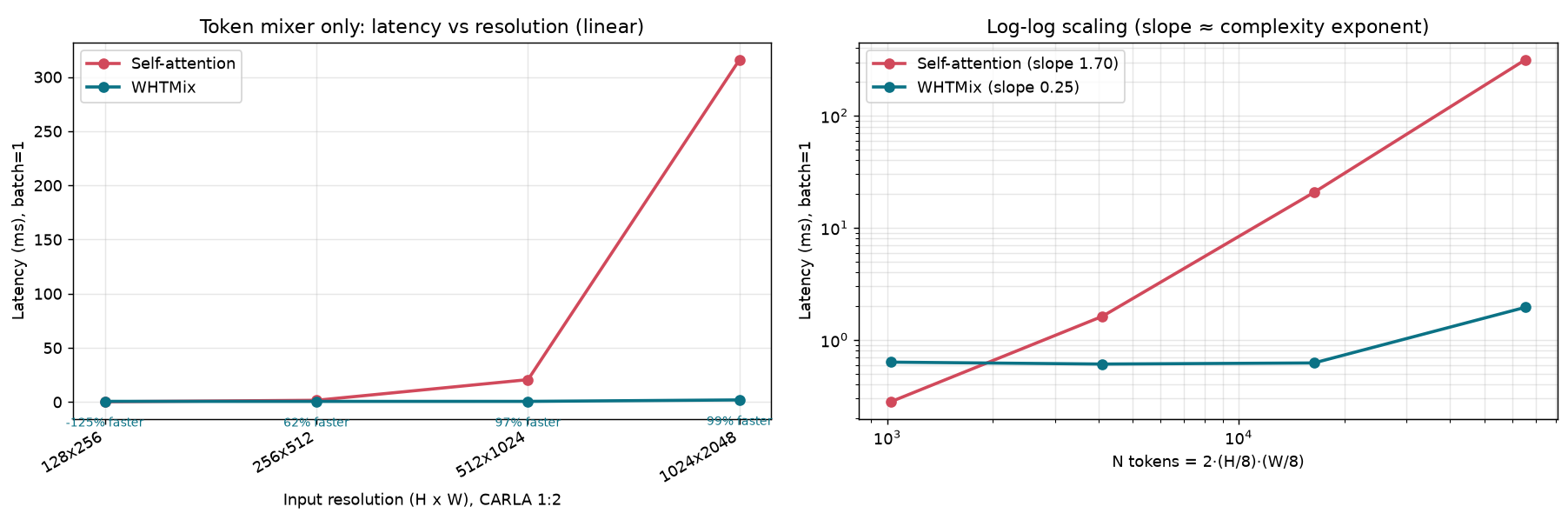}}
\caption{Latency of the token mixer in isolation: only the module that differs
between the two models is timed, at batch~1 on an RTX 4090, with power-of-two
$N$ so WHTMix needs no zero padding. \emph{Left}: linear axes; self-attention runs
away with resolution while WHTMix stays essentially flat. \emph{Right}: log-log
axes, whose fitted slopes separate the two regimes ($1.70$, near-quadratic, for
self-attention; $0.25$, near-flat, for WHTMix). At the deployed $N{=}16{,}384$ the
mixer is $33\times$ cheaper, rising to $162\times$ at $N{=}65$k, but its fixed
launch overhead leaves it slower than the fused attention kernel at $N{=}1024$.}
\label{fig:module}
\end{figure}

Table~\ref{tab:latency} and Fig.~\ref{fig:latency} sweep input resolution.
End-to-end, the WHTMix speedup grows with resolution toward the ceiling of
Eq.~\eqref{eq:ceiling}; both curves stay
super-linear because the shared cross-attention dominates. Isolating the token mixer
alone (Fig.~\ref{fig:module}) exposes the true asymptotics: near-quadratic for
attention, near-flat for WHTMix, a $33\times$ gap at the deployed $N{=}16{,}384$.

\begin{table}[H]
\caption{End-to-end batch-1 latency vs.\ input resolution. $N$ is the joint token count; speedup is baseline over WHTMix.}
\label{tab:latency}
\centering
\begin{tabular}{lrrrr}
\toprule
Resolution & $N$ & Baseline\,(ms) & WHTMix\,(ms) & Speedup \\
\midrule
$256\times512$   & 4{,}096  & 8.6   & 7.5   & $1.15\times$ \\
$384\times768$   & 9{,}216  & 33.6  & 14.8  & $2.27\times$ \\
$512\times1024$  & 16{,}384 & 95.6  & 36.1  & $2.65\times$ \\
$640\times1280$  & 25{,}600 & 226.3 & 80.9  & $2.80\times$ \\
$768\times1536$  & 36{,}864 & 453.5 & 166.5 & $2.72\times$ \\
$896\times1792$  & 50{,}176 & 848.8 & 298.0 & $2.85\times$ \\
\bottomrule
\end{tabular}
\end{table}

\subsection{Beyond Stereo: Long-Range Arena}
\begin{figure}[H]
\centerline{\includegraphics[width=0.3\textwidth]{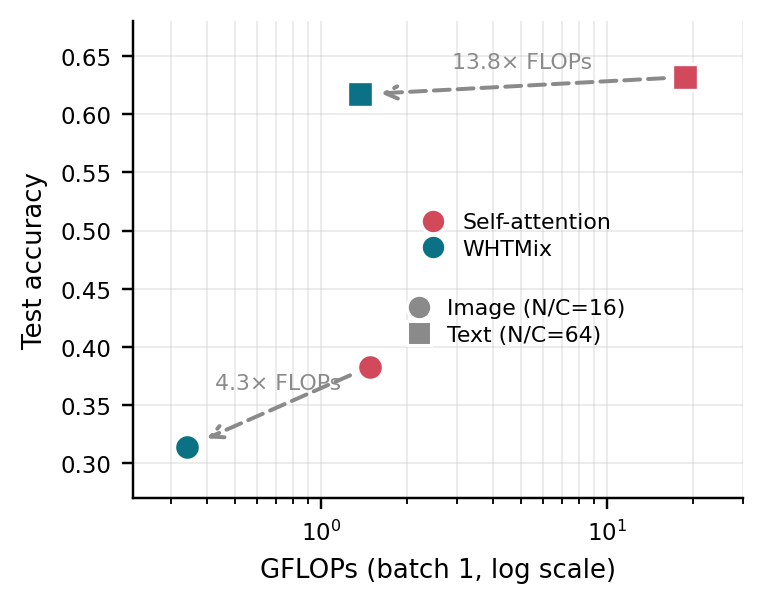}}
\caption{Test accuracy vs.\ batch-1 GFLOPs (log scale) on two non-stereo Long-Range
Arena tasks; color denotes the mixer and marker shape the task (circle: byte-level
image, $N{=}1024$, $N/C{=}16$; square: byte-level text, $N{=}4096$, $N/C{=}64$).
Each dashed arrow runs from self-attention to WHTMix within a task and is labelled
with the resulting FLOP reduction. At the higher $N/C$ the arrow is both longer and
flatter: more compute saved, less accuracy given up.}
\label{fig:lra}
\end{figure}

To test whether the benefit is stereo-specific or depends only on the
token-to-channel ratio, we insert WHTMix as a drop-in self-attention replacement
into a plain sequence classifier ($C{=}64$, $L{=}4$, $8$ heads) and evaluate on two
Long-Range Arena~\cite{lra} tasks unrelated to stereo: byte-level image
classification (CIFAR-10 grayscale, $N{=}1024$) and byte-level text sentiment
(IMDB, $N{=}4096$). Only the token mixer differs between paired runs.
Table~\ref{tab:lra} and Fig.~\ref{fig:lra} reproduce the stereo trend: the compute
reduction grows from $4.3\times$ at $N/C{=}16$ to $13.8\times$ at $N/C{=}64$, and
the wall-clock advantage is absent at the small size, where WHTMix is
launch-overhead-bound and slower than the fused attention kernel, emerges at the
larger one ($1.9\times$ faster). Accuracy follows the same regime: on the
long-sequence text task WHTMix is within $1.6$ points of attention, while on the
shorter image task the fixed mixer trails by $6.9$ points. The measured savings fall
short of the module-level $2+N/C$ of Eq.~\eqref{eq:speedup} ($18\times$ and
$66\times$), but by a similar factor in both tasks ($4.2\times$, $4.8\times$),
consistent with a fixed non-mixer remainder of embeddings, feed-forward and
classifier, the same Amdahl ceiling seen in stereo. Latency shows the mechanism:
WHTMix is essentially flat across the two tasks ($3.22\!\to\!3.14$~ms) while
attention grows $4.1\times$ ($1.47\!\to\!6.10$~ms). On unrelated data, then,
the mixer's utility is governed by $N/C$ rather than by the task; these are
single-seed runs and we defer multi-seed confirmation to future work.

\begin{table}[H]
\caption{Beyond stereo: WHTMix vs.\ self-attention on two Long-Range
Arena~\cite{lra} tasks (byte-level), identical config ($C{=}64$, $L{=}4$), single
seed; only the mixer differs. Accuracy, GFLOPs, and batch-1 latency on an RTX 4090.}
\label{tab:lra}
\centering
\setlength{\tabcolsep}{4pt}
\begin{tabular}{lccccc}
\toprule
Task ($N/C$) & Mixer & Acc. & GFLOPs & Lat.\,(ms) & FLOP$\downarrow$ \\
\midrule
Image (16) & attn   & \textbf{0.383} & 1.48  & \textbf{1.47} & --- \\
Image (16) & WHTMix & 0.314          & \textbf{0.34} & 3.22 & $4.3\times$ \\
\midrule
Text (64)  & attn   & \textbf{0.632} & 18.81 & 6.10 & --- \\
Text (64)  & WHTMix & 0.617          & \textbf{1.37} & \textbf{3.14} & $13.8\times$ \\
\bottomrule
\end{tabular}
\end{table}

\subsection{Discussion}
The results support a scoped claim: for high-resolution, low-channel transformers
($N\!\gg\!C$), dense-prediction vision in particular, replacing an $O(N^2)$ global
self-attention with an $O(N\log N)$ mixer yields large, resolution-growing compute
savings at parity accuracy, while genuine cross-attention is retained for
correspondence. The KITTI gap
indicates that a fixed mixer adapts less well than attention when real training data
are scarce and have sparse ground-truth, motivating larger-scale real-data with dense ground-truth pre-training as future work, along with applying the mixer to other dense-prediction transformers.

\section{Conclusion}
We replaced the joint self-attention of a stereo transformer with a learnable
Walsh-Hadamard token mixer, retaining data-dependent cross-attention for matching.
It matches attention accuracy on synthetic dense ground-truth data at $2.46\times$ less compute and
$2.65\times$ lower latency, a benefit our $2+N/C$ analysis both predicts and bounds.
Separately, a hybrid log-disparity loss reduces distant-object error at no latency
cost. Future work includes multi-seed, multi-model experimentation and larger real-data dense ground-truth pre-training.

\bibliographystyle{IEEEtran}
\bibliography{whtmix}

\end{document}